\documentclass{article}

\usepackage[english]{babel}		
\usepackage[utf8]{inputenc}
\usepackage{amsmath}			
\usepackage{amsfonts}
\usepackage{amssymb}
\usepackage{amsthm}
\usepackage{ifthen}
\usepackage{moreverb}			
\usepackage{color}

\usepackage{graphicx} 
\usepackage{subfigure} 

\usepackage{natbib}

\usepackage{algorithm}
\usepackage{algorithmic}

 \usepackage[accepted]{icml2013}

\newcommand{\field}[1]{\mathbb{#1}}

\newcommand{\R}{\field{R}}

\newcommand{\vect}[1]{\boldsymbol{#1}} 
\newcommand{\mat}[1]{\boldsymbol{#1}} 
\newcommand{\tvect}[1]{\tilde{\boldsymbol{#1}}} 
\newcommand{\tmat}[1]{\tilde{\boldsymbol{#1}}} 
\newcommand{\hvect}[1]{\hat{\boldsymbol{#1}}} 
\newcommand{\hmat}[1]{\hat{\boldsymbol{#1}}} 
\newcommand{\bvect}[1]{\bar{\boldsymbol{#1}}} 
\newcommand{\bmat}[1]{\bar{\boldsymbol{#1}}} 
\newcommand{\vzero}{\vect{0}}

\newcommand{\mxzero}{\mat{0}}

\newcommand{\dummystring}{QWERTYU}
\newcommand{\vci}[3][\dummystr]{\ifthenelse{\equal{#1}{\dummystring}}{\vect{#2}_{#3}}{\vect{#2}_{#3}^{(#1)}}}
\newcommand{\mx}[3][\dummystr]{\ifthenelse{\equal{#1}{\dummystring}}{\mat{#2}_{#3}}{\mat{#2}_{#3}^{(#1)}}}
\newcommand{\tvci}[3][\dummystr]{\ifthenelse{\equal{#1}{\dummystring}}{\tvect{#2}_{#3}}{\tvect{#2}_{#3}^{(#1)}}}
\newcommand{\tmx}[3][\dummystr]{\ifthenelse{\equal{#1}{\dummystring}}{\tmat{#2}_{#3}}{\tmat{#2}_{#3}^{(#1)}}}
\newcommand{\hvci}[3][\dummystr]{\ifthenelse{\equal{#1}{\dummystring}}{\hvect{#2}_{#3}}{\hvect{#2}_{#3}^{(#1)}}}
\newcommand{\hmx}[3][\dummystr]{\ifthenelse{\equal{#1}{\dummystring}}{\hmat{#2}_{#3}}{\hmat{#2}_{#3}^{(#1)}}}
\newcommand{\bvci}[3][\dummystr]{\ifthenelse{\equal{#1}{\dummystring}}{\bvect{#2}_{#3}}{\bvect{#2}_{#3}^{(#1)}}}
\newcommand{\bmx}[3][\dummystr]{\ifthenelse{\equal{#1}{\dummystring}}{\bmat{#2}_{#3}}{\bmat{#2}_{#3}^{(#1)}}}

\DeclareMathOperator{\trace}{tr}
\DeclareMathOperator{\diag}{diag}

\newcommand{\Ex}{\mathrm{E}}

\newcommand{\KL}{\mathrm{D}}

\newcommand{\Ent}{\mathrm{H}}

\newcommand{\Id}{\mat{I}}

\newcommand{\tabref}[1]{Table~\ref{tab:#1}}
\newcommand{\secref}[1]{Section~\ref{sec:#1}}
\renewcommand{\eqref}[1]{Eq.~\ref{eq:#1}}
\newcommand{\eqp}[1]{Eq.~\ref{eq:#1}}

\newcommand{\vd}[2][\dummystring]{\vci[#1]{d}{#2}}

\newcommand{\vg}[2][\dummystring]{\vci[#1]{g}{#2}}
\newcommand{\vh}[2][\dummystring]{\vci[#1]{h}{#2}}

\newcommand{\vm}[2][\dummystring]{\vci[#1]{m}{#2}}

\newcommand{\vv}[2][\dummystring]{\vci[#1]{v}{#2}}
\newcommand{\vw}[2][\dummystring]{\vci[#1]{w}{#2}}
\newcommand{\vx}[2][\dummystring]{\vci[#1]{x}{#2}}
\newcommand{\vy}[2][\dummystring]{\vci[#1]{y}{#2}}
\newcommand{\vz}[2][\dummystring]{\vci[#1]{z}{#2}}
\newcommand{\valpha}[2][\dummystring]{\vci[#1]{\alpha}{#2}}

\newcommand{\vth}[2][\dummystring]{\vci[#1]{\theta}{#2}}
\newcommand{\vmu}[2][\dummystring]{\vci[#1]{\mu}{#2}}

\newcommand{\veta}[2][\dummystring]{\vci[#1]{\eta}{#2}}

\newcommand{\vlam}[2][\dummystring]{\vci[#1]{\lambda}{#2}}

\newcommand{\vrho}[2][\dummystring]{\vci[#1]{\rho}{#2}}

\newcommand{\bvm}[2][\dummystring]{\bvci[#1]{m}{#2}}

\newcommand{\bvv}[2][\dummystring]{\bvci[#1]{v}{#2}}

\newcommand{\mxa}[2][\dummystring]{\mx[#1]{A}{#2}}

\newcommand{\mxi}[2][\dummystring]{\mx[#1]{I}{#2}}

\newcommand{\mxv}[2][\dummystring]{\mx[#1]{V}{#2}}
\newcommand{\mxw}[2][\dummystring]{\mx[#1]{W}{#2}}

\newcommand{\mxsigma}[2][\dummystring]{\mx[#1]{\Sigma}{#2}}

\newcommand{\srng}[2][1]{\{{#1},\dots,{#2}\}}

\newcommand{\gauss}{\mbox{${\cal N}$}}
\newcommand{\lse}{\mbox{lse}}
\newcommand{\half}{\mbox{$\frac{1}{2}$}}

\icmltitlerunning{Fast Dual Variational Inference for Non-Conjugate LGMs}

\begin{document} 

\twocolumn[
\icmltitle{Fast Dual Variational Inference for Non-Conjugate\\ Latent Gaussian Models}

\icmlauthor{Mohammad Emtiyaz Khan}{emtiyaz.khan@epfl.ch}
\icmladdress{School of Computer and Communication Sciences, Ecole Polytechnique F\'{e}d\'{e}rale de Lausanne, Switzerland}
\icmlauthor{Aleksandr Y. Aravkin}{saravkin@us.ibm.com}
\icmladdress{Numerical Analysis and Optimization, IBM T.J. Watson Research Center, Yorktown Heights, NY, USA}
\icmlauthor{Michael P. Friedlander}{mpf@cs.ubc.ca}
\icmladdress{Department of Computer Science, University of British Columbia, Vancouver, Canada}
\icmlauthor{Matthias Seeger}{matthias.seeger@epfl.ch}
\icmladdress{School of Computer and Communication Sciences, Ecole Polytechnique F\'{e}d\'{e}rale de Lausanne, Switzerland}

\icmlkeywords{Bayesian inference, variational inference, latent Gaussian models, variational Gaussian approximation}

\vskip 0.3in
]

\begin{abstract}
Latent Gaussian models (LGMs) are widely used in statistics and machine learning.
Bayesian inference in non-conjugate LGMs is difficult due to intractable integrals involving the Gaussian prior and non-conjugate likelihoods.
Algorithms based on variational Gaussian (VG) approximations are widely employed since they strike a favorable balance between accuracy, generality, speed, and ease of use.
However, the structure of the optimization problems associated with
these approximations  
remains poorly understood, and standard solvers take too long to converge.
We derive a novel dual variational inference approach that exploits the convexity property of the VG approximations. 
We obtain an algorithm that solves a convex optimization problem, reduces the number of variational parameters, 
and converges much faster than previous methods.
Using real-world data, we demonstrate these advantages on a variety of LGMs, including Gaussian process classification, and latent Gaussian Markov random fields.
\end{abstract}

\section{Introduction}\label{sec:intro}

Latent Gaussian models (LGM) are ubiquitous in machine learning and statistics
(e.g., Gaussian process models, Bayesian generalized linear models,
dynamical systems with non-Gaussian observations, robust PCA, and
non-conjugate matrix factorization). 
In many real-world applications, the likelihood is not conjugate to the Gaussian distribution, 
making exact Bayesian inference intractable.
These modern applications, especially those with large latent
dimensionality 
and number of observations, 
require fast, robust, and reliable algorithms for
approximate inference.

In this context, algorithms based on variational Gaussian (VG) approximations are growing in popularity
\cite{opper2009variational,challis2011concave,Lazaro:11,Honkela:11},
since they strike a favorable balance between accuracy, generality, speed, and ease of use.
However, compared to other approximations such as that of \citet{Seeger:11}, 
the structure of optimization problems associated with VG approximations  
remains poorly understood, and standard solvers for optimization take too long to converge.

While some variants of VG inference are convex \cite{Khan12nips}, they require $O(L^2)$
variational parameters to be optimized, where $L$ is the dimensionality of the latent Gaussian vector.
This slows down the optimization dramatically.
One approach is to restrict the covariance representations up front,
whether by naive mean field \cite{braun2010variational,knowles2011non} or
restricted Cholesky assumptions \cite{challis2011concave}.
Unfortunately, this can result in considerable loss in accuracy, since typical LGMs,
such as Gaussian processes, are tightly coupled.
Another approach is to reduce the number of parameters to $O(N)$, where $N$ 
is the dimension of the observation vector, 
using an exact covariance parameterization \cite{opper2009variational}.
This reparameterization destroys the convexity of the original problem, and
very slow convergence is typically observed \cite{Khan12nips}.
 A recent
coordinate-ascent method improves upon the state of the art \cite{Khan12nips},
but is restricted to Gaussian process models only and uses inefficient
low-rank matrix updates.

We propose a dual decomposition approach that allows us to 
reduce the number of parameters to $O(N)$ while retaining convexity.
The new dual optimization problem can be solved very rapidly with standard methods for smooth optimization. 
Using real-world data, we demonstrate that our algorithm converges much faster than the state of the art on a variety of LGMs.
Unlike the approach of \citet{Khan12nips}, our algorithm is generic and is not restricted to Gaussian processes.

\section{Latent Gaussian Models}\label{sec:model}
Given a vector of observations $\vy{}\in{\cal Y}^N$, the dependencies among its components can be modeled 
using a latent vector $\vz{}\in\R^L$. 
Here, the set $\cal Y$ is the domain of each observation, e.g., for binary observations, ${\cal Y} = \{ 0,1 \}$.
The latent vector $\vz{}$ is assumed to follow a Gaussian distribution 
$p(\vz{}) = \gauss(\vz{}|\vmu{},\mxsigma{})$. 
The likelihood has the general form
\begin{align}
  p(\vy{}|\vz{}) &= \prod_{n=1}^N p(y_n|\eta_n),\quad \veta{}=\mxw{}\vz{},
\end{align}
where  $\mxw{}\in\R^{N\times L}$.
Model parameters $\vth{}$ consist of parameters required to specify $\vmu{}$, $\mxsigma{}$, $\mxw{}$, as well
as parameters of the distribution $p(y_n|\eta_n)$. All densities are implicitly
conditioned on $\vth{}$, which we suppress from the notation.
Also note that $\eta_n$ can be a vector but we restrict ourselves to scalar $\eta_n$.
Our results can be easily extended to the vector case.

\begin{figure}[!t]
\centering
\subfigure[]{\includegraphics[height = 2in]{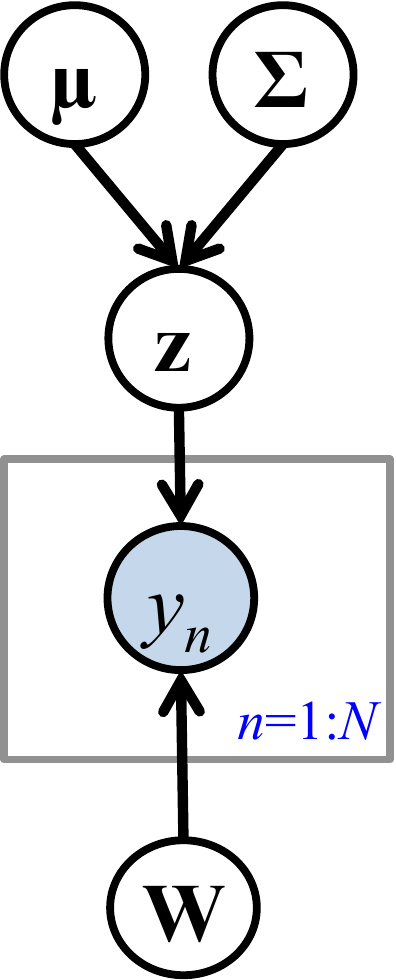}
\label{fig:LGMa}}
\hspace{.2in}
\subfigure[]{\includegraphics[height = 2in]{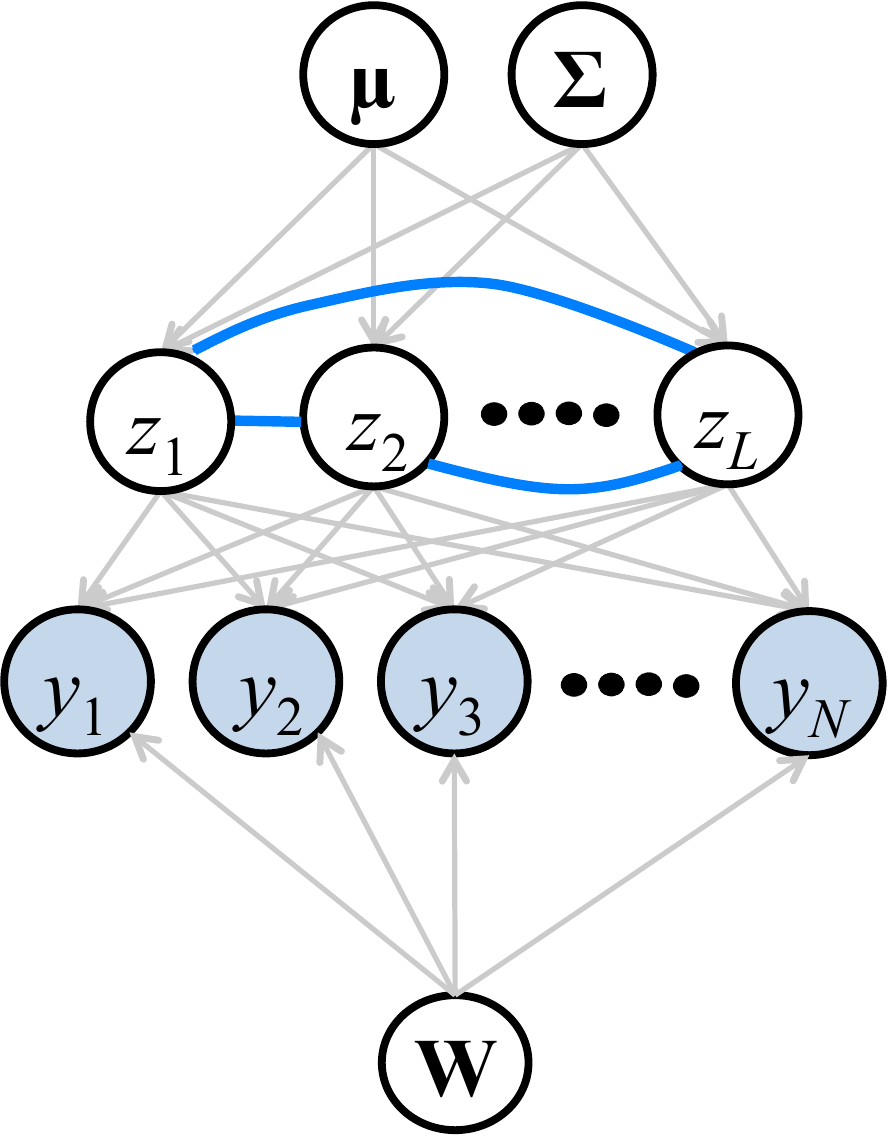}
\label{fig:LGMb}}
\caption{The graphical model for latent Gaussian models shown in left figure, and expanded in the right figure to explicitly show the correlation in the latent vector $\vz{}$ induced due to a non-diagonal $\mxsigma{}$.}
\end{figure}

Many models used in statistics and machine learning are instances of LGMs.
Several examples are listed in Table \ref{tab:examplesLGM}, and an extensive list can be found in \citet[Chapter 1]{emtThesis}.
Bayesian generalized linear models constitute one such example, where we assume 
a latent Gaussian weight vector and  use exponential
family likelihoods with natural parameter $\eta_n$.
Similarly, latent Gaussian Markov random fields (GMRF) model spatial correlations 
by using a GMRF with a sparse inverse covariance matrix $\mxsigma{}^{-1}$, 
along with an exponential family likelihood to model non-normal observations \cite{Rue05}.
For example, count data with spatial dependence (e.g., incidences
of a disease in different regions of a country) can be modeled using a Poisson likelihood with rate $r_n = \exp(\eta_n)$.
The log-Gaussian Cox process is a non-parameteric generalization of this
setting \cite{Rue09}.
Other non-parameteric examples are  Gaussian process (GP) models,
where observation pairs $\{y_n,\vx{n}\}$ are modelled via a latent Gaussian
process $z(\vx{})$ with the prior specified by mean and covariance functions.

In Bayesian inference, we wish to compute expectations with respect to the posterior distribution
\begin{align}
  p(\vz{}|\vy{}) &\propto \prod\nolimits_{n=1}^N p(y_n|\eta_n)
  \gauss(\vz{}|\vmu{},\mxsigma{}). \label{eq:bayes-infer}
\end{align}
For example, prediction of a new observation $y_*$ can be obtained by computing the expectation
$p(y_*|\vy{}) = \int p(y_*|\veta{}) p(\vz{}|\vy{})\, d\vz{}$.
Another important task is computation of the marginal likelihood
\begin{align}
  p(\vy{}) &=\int \prod\nolimits_{n=1}^N p(y_n|\eta_n)
  \gauss(\vz{}|\vmu{},\mxsigma{})\, d\vz{}.
\end{align}
For example, parameters $\vth{}$ can be learned by maximizing the log of the marginal likelihood, 
$\log p(\vy{})$.
This is also referred to as empirical Bayes or automatic relevance determination (ARD) \cite{Tipping:01, Rasmussen06}.

For non-Gaussian likelihoods, both of these tasks are intractable.
Applications in practice demand good approximations that scale favorably in $N$ and $L$.

\begin{table*}[!t]
\begin{center}
\begin{tabular}{|l||c|c|l|l|l|c|}
\hline
Model  & Data & $\vz{}$ & $\vth{}$ & $N$ & $L$ & Remarks\\	
\hline
\hline
Bayesian Logistic & $\{y_n,\vx{n}\}$ & Regression weights & ${\vmu{},\mxsigma{}}$& \#Obs & \#Features & Row of $\mxw{}$\\
Regression &  & $y_n\leftarrow f(\vz{}^T\vx{n})$ & & & & $\vw{n} = \vx{n}$ \\	
\hline
Gaussian Process & $\{y_n,\vx{n}\}$ &Regression function &${s,\sigma}$& \#Obs &\#Features & $\mxw{} = \mxi{}$ \\
Classification &  & $y_n\leftarrow f(z_n)$ 		& & & & $N = L$ \\	
\hline
Gaussian Markov & $\{y_{n}\}$ & Latent Gaussian field & ${k_v, k_u}$ & \#Obs & \# Latent & \\	
Random Field & & $y_n \leftarrow f(\vz{n})$  & & & dims & \\
\hline
Probabilistic PCA & $\{y_{ni}\}$ & Latent factors & $\mxw{}$ & \#Obs & \#Latent & $N > L$\\	
 & & $y_{n}\leftarrow f(\vw{i}^T\vz{n})$  & & dims & factors & $\vmu{}=\vzero, \mxsigma{} = \mxi{}$\\
\hline
\end{tabular}
\end{center}
\caption{Examples of LGM. Each column is a quantity from our generic
  LGM definition. Each row shows corresponding quantities for a
  model. First two models are supervised and the last two are
  unsupervised. For columns 2 and 3, $n$ ranges over $1$ to $N$ and
  $\{a_{n}\}$ denotes the set of variables indexed by all values of
  $n$. $y\leftarrow f(z)$ implies that $y$ can be generated using some
  function $f$ of $z$. In last three columns, `Obs' means
  observations, `Dims' means dimensions, and `\#' represents the
  number of a quantity. For GP, $s$ and $\sigma$ are hyperparameters
  of the covariance function. Similarly, $k_u$ and $k_v$ are
  hyperparameters for the latent field. See Section \ref{sec:exper}
  for details. For PPCA, the subscript $i$ indexes the observation vector.}
\label{tab:examplesLGM}
\end{table*}


\section{Variational Gaussian Inference} \label{sec:infer}
In the {\em variational Gaussian} approximation \cite{opper2009variational}, we assume the posterior
to be a Gaussian $q(\vz{}) = \gauss(\vz{}|\vm{},\mxv{})$.
The posterior mean $\vm{}$ and covariance $\mxv{}$ form the set of variational parameters, and are chosen to maximize 
the variational lower bound to the log marginal likelihood shown in Eq.~\ref{eq:LB}.
To get this lower bound, we first multiply and divide by $q(\vz{})$ in Eq.~\ref{eq:LB0},
 and then use Jensen's inequality and the concavity of $\log$
(we denote the expectation with respect to $q(\vz{})$ by $\Ex_{q(z)}$):
\begin{align}
   \log p(\vy{}) &= \log\int q(\vz{})\frac{\prod_n p(y_n|\eta_n) p(\vz{})}{q(\vz{})}\, d\vz{} 
	\label{eq:LB0}\\
  & \ge \Ex_{q(z)}\left[ \log\frac{\prod_n p(y_n|\eta_n) p(\vz{})}{q(\vz{})}
	\label{eq:LB}
  \right].
\end{align}
The lower bound can be simplified further, and variational parameters $\vm{}$ and $\mxv{}$ 
can be obtained by maximizing it:
\begin{equation}\label{eq:orig-problem}
  \max_{\vm{},\mxv{}\succ
\mxzero} -\KL[q(\vz{})\, \|\, p(\vz{})] - \sum_{n=1}^N
  \Ex_{q(\eta_n)}[-\log p(y_n|\eta_n)],
\end{equation}
where 
\begin{align}
\KL[q\, \|\, p] &= \Ex_q[\log q(\vz{}) - \log p(\vz{})] \\
q(\eta_n) &= \gauss(\bar{m}_n,\bar{v}_n) \\
\bvm{} &= \mxw{}\vm{}, \quad \bvv{} =\diag(\mxw{}\mxv{}\mxw{}^T).
\end{align}
See Eqs.~4--7 in \citet{Khan12aistats} for details of this derivation.

The first term in Eq.~\ref{eq:orig-problem} is the relative entropy, and is jointly concave in
$(\vm{},\mxv{})$.
The second term $\Ex_{q(\eta_n)}[-\log p(y_n|\eta_n)]$ is not always available in closed form. 
We assume in this paper that, in such cases, we can evaluate an upper
bound $f_n$ to this term, i.e.,
\begin{align}
  \Ex_{q(\eta_n)}[-\log p(y_n|\eta_n)] \le f_n(\bar{m}_n,\bar{v}_n).
\end{align}
This is also known as the local variational bound (LVB). 
We assume that $f_n$ is differentiable and---most importantly---convex.
We discuss a few such LVBs in \secref{details};
see \citet{emtThesis} for an extensive list.

The resulting optimization problem is shown below in Eq.~\ref{eq:primal-problem}
 and is expanded in Eq.~\ref{eq:primal-detail}:
\begin{align}
\label{eq:primal-problem}
& \max_{\vm{},\mxv{}\succ
\mxzero} -\KL[q(\vz{})\, \|\, p(\vz{})] - \sum_{n=1}^N f_n(\bar{m}_n,\bar{v}_n) \\\
\label{eq:primal-detail}
  &:= \half[\log|\mxv{}| - \trace(\mxv{}\mxsigma{}^{-1}) - (\vm{} -\vmu{})^T \mxsigma{}^{-1} (\vm{} -\vmu{})] \nonumber\\
	&\qquad\qquad\qquad\qquad - \sum_{n=1}^N f_n(\bar{m}_n,\bar{v}_n) + \textrm{cnst}.
\end{align}
The above lower bound is strictly concave \cite{braun2010variational, challis2011concave, emtThesis}.

\subsection{Related Work} \label{sec:relatedWork}
A straight-forward approach is to solve Eq. \ref{eq:primal-problem} directly in
$(\vm{},\mxv{})$ \cite{braun2010variational,challis2011concave,
marlin2011piecewise,Khan12aistats}. In practice, direct methods are slow and
memory-intensive because of the very large number $L + L(L+1)/2$ of primal variables.
\citet{challis2011concave} show that for log-concave
likelihoods $p(y_n|\eta_n)$, the original problem \eqp{orig-problem}
is jointly concave in $\vm{}$ and the Cholesky factor of $\mxv{}$, and
additional LVBs are not required.  This fact, however, does not result
in any reduction in number of parameters, and they propose to use
factorizations of a restricted form, which negatively affects the
approximation accuracy.

\citet{opper2009variational} and \citet{nickisch2008approximations} note
that the optimal $\mxv{*}$ must be of the form
\begin{align}
\label{eq:vstarOpperArch}
\mxv{*}=(\mxsigma{}^{-1} + \mxw{}^T(\diag\vlam{})\mxw{})^{-1},
\end{align}
which suggests reparameterizing
\eqp{primal-problem} in terms of $L+N$ parameters $(\vm{},\vlam{})$, 
where $\vlam{}$ is the new variable. However,
the problem is non-concave in this alternative parameterization \cite{Khan12nips}. 
Moreover, as shown in
\cite{Khan12nips} and our experiments here, convergence can be exceedingly
slow. The coordinate-ascent algorithm proposed
in \cite{Khan12nips} solves the problem of convergence, 
but seems limited to the case $N=L$ and $\mxw{}=\Id$. 
In addition, it requires $N$ rank-one updates of $\mxv{}$ per iteration, which is slow on
modern architectures optimized for block-matrix computations.

A range of different deterministic inference approximations apply to
latent Gaussian models. The local variational method is convex for log-concave
potentials and can be solved at very large scales \cite{Seeger:11}. However, it
applies to super-Gaussian\footnote{
  Neither the Poisson, nor the stochastic volatility likelihood are
  super-Gaussian (\secref{details}).}
potentials only. The bound it maximizes is provably less tight than
\eqp{orig-problem}
\cite{Seeger:09c,challis2011concave}, and it leads to worse results than the
variational Gaussian approximation in general \cite{nickisch2008approximations,
emtThesis}. A key interpretation of  this method is that it can be seen as one way to
generate LVBs (for super-Gaussian potentials), which can be used in our VG
setup \cite{Seeger:09c}. Expectation propagation \cite{Minka:01a,Seeger:07d} is
more general
and can be more accurate than most other approximations mentioned here.
Based on a saddlepoint rather than an optimization problem, the standard EP
algorithm does not always converge and can be numerically unstable. Among these alternatives, the variational Gaussian
approximation stands out as a compromise between accuracy and good algorithmic
properties, which is widely used beyond latent Gaussian model applications as
well \cite{Lazaro:11,Honkela:11}.

\section{Dual Variational Inference}\label{sec:decouple}
In this section, we show how \eqp{primal-problem} can be solved using 
a convex dual formulation in only $N$ variational parameters. As shown in our experiments,
the novel formulation admits simple algorithms which converge much more rapidly
and have a lower per-iteration cost than previous methods reviewed above.
We achieve this by dual decomposition: decoupling the two terms in \eqp{primal-problem} by
equality constraints, and then forming the Lagrangian dual.
To be precise, we first introduce two new variables $h_n, \rho_n \in \mathbb{R}$ for each $n$ 
and introduce constraints $h_n = \bar{m}_n$ and $\rho_n = \bar{v}_n$.
The resulting (equivalent) optimization problem can be written as
\begin{align}
  & \max_{\vm{},\mxv{},\vh{},\vrho{}} -\KL[q(\vz{})\, \|\, p(\vz{})] -
  \sum\nolimits_{n=1}^N f_n(h_n,\rho_n)  \\
  & \qquad\qquad \text{s.t.}\;\; \vh{}=\mxw{}\vm{},\quad
  \vrho{} = \diag(\mxw{}\mxv{}\mxw{}^T). \nonumber
\end{align}
Next, we introduce dual variables $\valpha{}, \vlam{}\in\R^N$ associated to these
constraints, and form the corresponding Lagrangian 
\begin{align}
  & {\cal L} = -\KL[q(\vz{})\, \|\, p(\vz{})] -
  \sum\nolimits_{n=1}^N f_n(h_n,\rho_n)\label{eq:lagrange}
 \\
  &\qquad + \valpha{}^T(\vh{}-\mxw{}\vm{}) +
  {\textstyle\frac{1}2}\vlam{}^T( \vrho{} - \diag(\mxw{}\mxv{}\mxw{}^T) ).\nonumber
   \end{align}
Strong duality holds because the constraints are affine, 
and so the solution to the original problem
can be found by minimizing the Lagrangian dual with respect to
$(\valpha{}, \vlam{})$, i.e.,
\begin{align}
 \min_{\valpha{},\vlam{}}\ \mathcal{D}(\valpha{}, \vlam{})= \min_{\valpha{},\vlam{}} \max_{\vm{},\mxv{},\vh{},\vrho{}} {\cal L}. \label{eq:decoupledDual} 
\end{align}

The advantage of this formulation is that we
can solve analytically for $(\vm{},\mxv{})$ and $(\vh{},\vrho{})$,
 and the resulting dual $\mathcal{D}(\valpha{},\vlam{})$ is available in closed form.
Since $\valpha{}$ and $\vlam{}$ are length $N$ vector, the dual minimization involves only $O(N)$ parameters.
 
Derivations of the following statements are given in the Appendix. 
The unique maximizer with respect to $(\vm{},\mxv{})$ is given by
\begin{align}
\label{eq:mstar}
 	\vm{*} &= \vmu{} - \mxsigma{}\mxw{}^T\valpha{}\\
\label{eq:Vstar}
	\mxv{*} &= \mxa{\vlam{}}^{-1} := (\mxsigma{}^{-1} + \mxw{}^T(\diag\vlam{})\mxw{})^{-1}.
\end{align}
Importantly, $\mxv{*}$ has precisely the economical form pointed out by
\citet{opper2009variational}. 

Maximization over $(\vh{},\vrho{})$ is also available in closed form.
Collecting the terms involving $(h_n,\rho_n)$ in Eq.~\ref{eq:lagrange},
 we get the following optimization problem, 
\begin{equation}\label{eq:conjugate}
  f_n^*(\alpha_n,\lambda_n) := \max_{h_n,\rho_n} \alpha_n h_n + \lambda_n \rho_n/2
  - f_n(h_n,\rho_n),
\end{equation}
which is in fact the $f_n^*$ the {\em Fenchel conjugate} of $f_n$ \cite{Rockafellar:70}, and is 
convex and well-defined due to the convexity of $f_n$.
For many likelihoods (and LVBs), $f_n^*$ is available in closed form.
We give several examples in \secref{details}, summarized in Table \ref{tab:fenchelDual}.

Note that the effective domain of $f_n^*$ (i.e., values of $(\alpha,\lambda)$ for which $f_n^*$ is finite) may be restricted.  
We give details of this and show the effective domain of $f_n^*(\alpha_n,\lambda_n)$ for several commonly used likelihoods in
\secref{details}.
We denote the effective domain of $f_n^*$ by $\mathcal{S}$.

Plugging in Eq.~\ref{eq:mstar}, \ref{eq:Vstar}, and \ref{eq:conjugate} into Eq.~\ref{eq:lagrange} and ignoring the constants,
directly gives us the optimization problem
\begin{align}\label{eq:dual-problem}
  \min_{\valpha{},\vlam{}\in\mathcal{S}} 
	\half \valpha{}^T\widetilde{\mxsigma{}}\valpha{}
	- \widetilde{\vmu{}}^T\valpha{}
	 - \half\log|\mxa{\vlam{}}|
	 + \sum_{n=1}^N f_n^*(\alpha_n,\lambda_n),
\end{align}
where $\widetilde{\vmu{}} = \mxw{}\vmu{}$ and $\widetilde{\mxsigma{}} = \mxw{}\mxsigma{}\mxw{}^T$.

This is a strictly convex optimization problem involving $2N$ parameters, in contrast to 
\eqp{primal-problem}, which involves $O(L^2)$ number of parameters.
Given $(\valpha{*},\vlam{*})$ that minimizes the dual, the primal solution $(\vm{*},\mxv{*})$ 
is obtained using Eq.~\ref{eq:mstar} and \ref{eq:Vstar}.
It might appear that minimizing the dual might be a difficult problem due to the constraints, but
as we show later $f_n^*$, act as barrier functions, which simplify the optimization.

\comment{
 For
likelihoods discussed in this paper, it turns out that $\valpha{}$ can be
eliminated, and we need to optimize over $\vlam{}$ only (for which further
constraints, such as $\lambda_n>0$, apply). Details
are given in the next section, where we also present a highly efficient
algorithm for solving the dual \eqp{dual-problem}.
}

\comment{
Define
\begin{equation}\label{eq:r-func}
  R(\valpha{}) :=
  {\textstyle\frac{1}2}\valpha{}^T\mxw{}\mxsigma{}\mxw{}^T\valpha{} -
  \vmu{}^T\mxw{}^T\valpha{}.
\end{equation}
Plugging $(\vm{*},\mxv{*})$ into \eqp{lagrange}, the following dual problem
is obtained (up to an additive constant):
\begin{align}
  \min_{\valpha{},\vlam{}}\max_{\vh{},\vrho{}}\; & R(\valpha{}) -
  {\textstyle\frac{1}2}\log|\mxa{}| \label{eq:dual-first} \\
  & + \sum\nolimits_{n=1}^N \alpha_n h_n + \lambda_n \rho_n/2 - f_n(h_n,\rho_n).
  \nonumber
\end{align}

Plugging this in, we arrive at the final form of the dual problem for
\eqp{primal-problem}:
\begin{equation}\label{eq:dual-problem}
  \min_{\valpha{},\vlam{}} R(\valpha{}) -
  {\textstyle\frac{1}2}\log|\mxa{}(\vlam{})| + \sum\nolimits_{n=1}^N
  f_n^*(\alpha_n,\lambda_n).
\end{equation}
}

\begin{table*}[t]
\centering
\begin{tabular}{|l||l|l|l|l|}
\hline
 & Poisson & Bernoulli-logit & Multi-Logit & Stochastic Volatility \\
\hline
\hline
$p(y|\eta)$ & $\exp(y\eta - e^{\eta})/y!$ & $e^{y\eta}/(1+e^\eta)$ & $\exp(\vy{}^T\veta{})/\sum_k \exp(\eta_k)$ & $\gauss(y|0, e^\eta)$ \\ 
\hline
LVB & Not required & Yes & Yes & Not required \\
\hline
$f(h,\rho)$ & $-yh + e^{h+\rho/2}$ & $-yh + \log(1+ e^{h + \rho/2})$ & $-\vy{}^T\vh{} + \lse(\vh{} + \half\vrho{})$ & $\half h + \half y^2 e^{-h+\rho/2}$ \\
\hline
$f^*(\lambda)$ & $\lambda(\log\lambda -1)$ & $\lambda\log\lambda $ & $\sum_{k=1}^{K-1} \lambda_k \log \lambda_k + t\log t$ & $\lambda \log(2\lambda/y^2) - \lambda$\\ 
& & $ + (1-\lambda)\log(1-\lambda)$ & where $t := \sum_{k=1}^{K-1} \lambda_k$ & \\
\hline
Range $\mathcal{S}$ & $\lambda>0$ & $\lambda\in(0,1)$ & $\lambda_k>0,\, t < 1$ & $\lambda>0$ \\
\hline
\end{tabular}
\caption{This table summarizes LVBs (or exact expressions) and Fenchel conjugates for a number of likelihoods. Stochastic volatility is from \cite{Rue09}, the Bernoulli-logit  multi-logit bound from \cite{Blei06}. Here, $\lse(\vv{}) = \log(1 + \sum_{k=1}^{K-1} e^{v_k})$. For first 3 columns, $\alpha$ is constrained to be equal to $\lambda-y$, and for the last one $\alpha = \half - \lambda$.}
\label{tab:fenchelDual}
\end{table*}

\section{Algorithmic Details}\label{sec:details}
Here we give details on the function $f_n$ and its conjugate
$f_n^*$.
We also provide computational details about our 
algorithm for solving the dual problem \eqp{dual-problem}.

\subsection{Fenchel conjugates}
We give an illustrative example to show the derivation of Fenchel conjugates.
For simplicity, we drop the subscript $n$.
Consider the Poisson likelihood
$\log p(y|\eta) = y\eta - \exp(\eta) + \textrm{cnst}$:
\begin{align}
f(h,\rho) = \Ex[-\log p(y|\eta)]  
  = -y h + e^{h + \rho/2} + \textrm{cnst}
\end{align}
This function is convex.
To determine the Fenchel conjugate $f^*(\alpha,\lambda)$, we use \eqp{conjugate} and
first maximize over $\rho$, obtaining
 $\lambda = e^{h+\rho_*/2}$. This implies $\lambda>0$, 
 since otherwise the conjugate takes the value $+\infty$.
Then,
\begin{align}
  f^*(\alpha,\lambda) & = \max_h \lambda(\log\lambda-1) + (\alpha+y-\lambda) h \\
  & = \lambda(\log\lambda-1) + \delta_{0}(\alpha - \lambda+y),
\end{align}
where $\delta_{0}(\cdot)$ is the convex indicator function, which equals 
$0$ if the argument is $0$, and $+\infty$ otherwise; the indicator term 
enforces the constraint $\alpha = \lambda - y$. 
Note that $\lambda$
is constrained to lie in $S=\{\lambda>0\}$.

 Examples of $f_n$ and $f_n^*$ for a range of other likelihood functions are given in
\tabref{fenchelDual}. Detailed derivation of these is available in an online appendix to the paper.
In all the cases, $\alpha_n=\lambda_n - y_n$ applies, except for the stochastic volatility where $\alpha_n = \half - \lambda_n$.

\subsection{Reduced dual}
As discussed in previous section, for all likelihoods discussed in this paper, we have a restriction on $\valpha{}$.
For example, for the first three likelihoods $\valpha{}= \vlam{} - \vy{}$.
Plugging this in Eq. \ref{eq:dual-problem}, we get the reduced dual
\begin{align}\label{eq:dual-simple}
&  \min_{\vlam{}\in\mathcal{S}} \
	\half (\vlam{} - \vy{})^T\widetilde{\mxsigma{}}(\vlam{} - \vy{})
	- \widetilde{\vmu{}}^T(\vlam{} - \vy{})
	 - \half\log|\mxa{\vlam{}}| \nonumber \\
&\qquad\qquad	 + \sum_{n=1}^N f_n^*(\lambda_n).
\end{align}
In other words, the
equality constraints $\alpha_n=\lambda_n - y_n$ are enforced by the domain of
the conjugate $f_n^*(\alpha_n,\lambda_n)$, which allows us to eliminate
$\valpha{}$ altogether using an affine substitution. 

\subsection{Algorithm Details}
In this section, we show that the constrained problem of Eq.~\ref{eq:dual-simple}
can be optimized efficiently using quasi-Newton methods. 
We make use of the fact that the Fenchel conjugates act as barrier functions,
thereby allowing us to limit the line search within the feasible set.
This way, we avoid any unnecessary function evaluations to get an efficient implementation,
treating the problem as if it was unconstrained.

First of all, note that the gradient of Eq.~\ref{eq:dual-simple} with
respect to $\vlam{}$ is given by
\begin{align}
\widetilde{\mxsigma{}}(\vlam{} -\vy{}) - \widetilde{\vmu{}}
-\half \diag(\mxw{}^T \mxa{\vlam{}}^{-1}\mxw{}) + \vg{\vlam{}}^*,
\end{align}
where $\vg{\vlam{}}^*$ is the vector of gradients of $f_n^*$ with respect to $\lambda_n$.
This gradient is used to obtain a descent descent direction $\vd{}$.

Given the descent direction $\vd{}$ and an initial step size $\delta_0$,
our goal is find a new step size $\delta$ while keeping $\vlam{}$ feasible.
We do this by restricting the linesearch to the feasible set only,
and then using Armijo or Wolfe condition in exactly the same way as in the unconstrained case.
We illustrate this for the constraint $\lambda_n>0$, which arise when Fenchel conjugate contains terms such as $\log(\lambda_n)$.
Other constraints can be implemented in a similar way.
Assume that the current $\vlam{}$ is in the feasible set, i.e., $\lambda_n >0$ for all $n$.
We find the indices $\mathcal{I}$ where $\lambda_n + \delta_0 d_n < 0$.
Since $\lambda_n >0$, we have $d_n<0$ for all $n\in \mathcal{I}$.
To keep the next $\lambda_n>0$, the largest step should be less than the minimum $\lambda_n/|d_n|$ of all $n\in\mathcal{I}$.
Hence, we restrict the search to the set
\begin{align}
\delta = (1-\epsilon) \min \left\{\min_{n\in \mathcal{I}} \frac{\lambda_n}{|d_n|}, \delta_0\right\}\;,
\end{align}
where $\epsilon>0$ ensures strict feasibility. 
Other constraints can be dealt with in a similar way.

\comment{
\begin{figure}[!t]
\center
\includegraphics[width=.8\columnwidth]{poissonConjugate-crop.pdf}
\label{fig:poissonConjugate}
\caption{MISSING HERE: Explanation, or kick out. Change notation $m\to h$, $v\to \rho$. WHAT IS $y$? \\
This figure shows that the Fenchel conjugate for poisson likelihood is finite only for $\alpha>0$. The function $h\mapsto f(h,1)$ is shown in blue (here, $\rho=1$). We choose $\alpha = 10$, and plot $\alpha h$ (solid red) and $-\alpha h$ (dashed black). The maximum gap is shown with cyan-dashed vertical lines. For positive slope, the gap is maximum at a finite $h$, while for negative slope, the gap is maximized at $-\infty$ with the value being $+\infty$.}
\end{figure}
}


\section{Experiments} \label{sec:exper}

In this section, we apply our novel dual variational algorithm to
a range of real-world Bayesian inference problems.
We compare our algorithm to the widely
used method of \citet{opper2009variational}, which plugs the covariance
parameterization of \ref{eq:vstarOpperArch} into the primal problem \eqp{primal-problem} and optimizes it over $(\vm{},\vlam{})$.
We refer to this method as `Opper-Arch'.
We do not present results for the naive method of solving the primal in $(\vm{},\mxv{})$ directly, since this turns out to be much slower than the alternatives.

\subsection{Multi-Way GP Classification}
In this section, we consider a multinomial logit $K$-way Gaussian process
classification (mGPC) model, following the experimental setup outlined in
\citet{Khan12aistats} and \cite{Girolami06}.
See \citet[Chapter 1]{emtThesis} for details how GP classification can be formulated as an LGM. 

For multinomial logit likelihood, the term $f_n$ is not available in closed form, and we use the LVB proposed by \citet{braun2010variational}.
Details of this LVB and its Fenchel conjugate are given in \tabref{fenchelDual}.

\comment{
A likelihood factor $p(y_n|\veta{n})$, $y_n\in\srng{K}$, is
parameterized by $\veta{n}\in\R^K$ (see \tabref{fenchelDual}), moreover
$\mxw{}=\Id$ and $\veta{}=[\veta{n}]=\vz{}$. We employ $K$ GPs
$z^{(k)}(\cdot)$, which are a priori independent, so that
\[
  p(\vz{}) = \prod_{k=1}^K\gauss( \vz[k]{} | \vzero,\mxsigma{} ),
\]
where $\mxsigma{}\in\R^{N\times N}$ is the kernel matrix evaluated at $N$
training input points. Note that in this section, we deviate from the notation
in the rest of the paper, in that $\vz{}\in\R^{N K}$ (since $L=N$), and
$\mxsigma{}$ is replaced by $\Id_K\otimes\mxsigma{}$. We use a Gaussian
posterior approximation which factorizes across classes: $q(\vz{}) =
\prod_{k=1}^K \gauss(\vz[k]{} | \vm[k]{},\mxv[k]{})$.}

\begin{figure*}[t]
\centering
\subfigure[]{\includegraphics[height=3in]{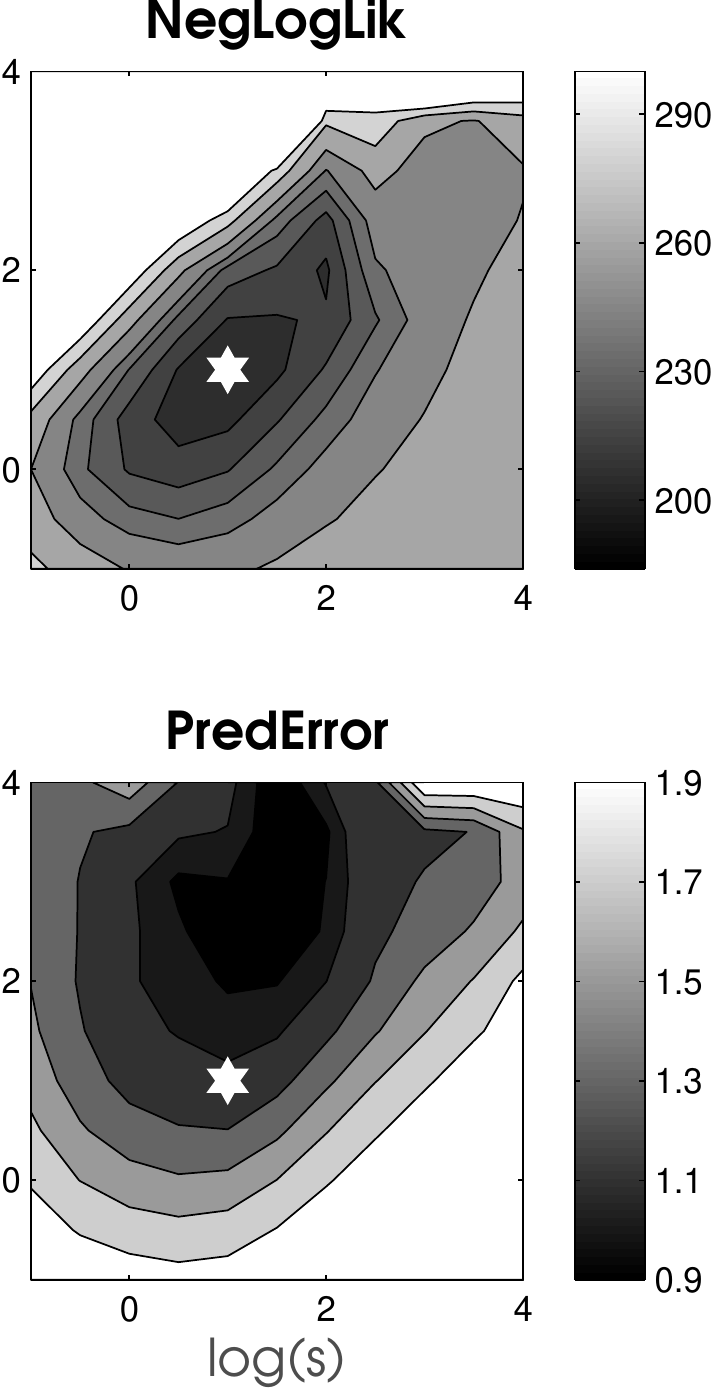}}
\hspace{.2in}
\subfigure[]{\includegraphics[height=3in]{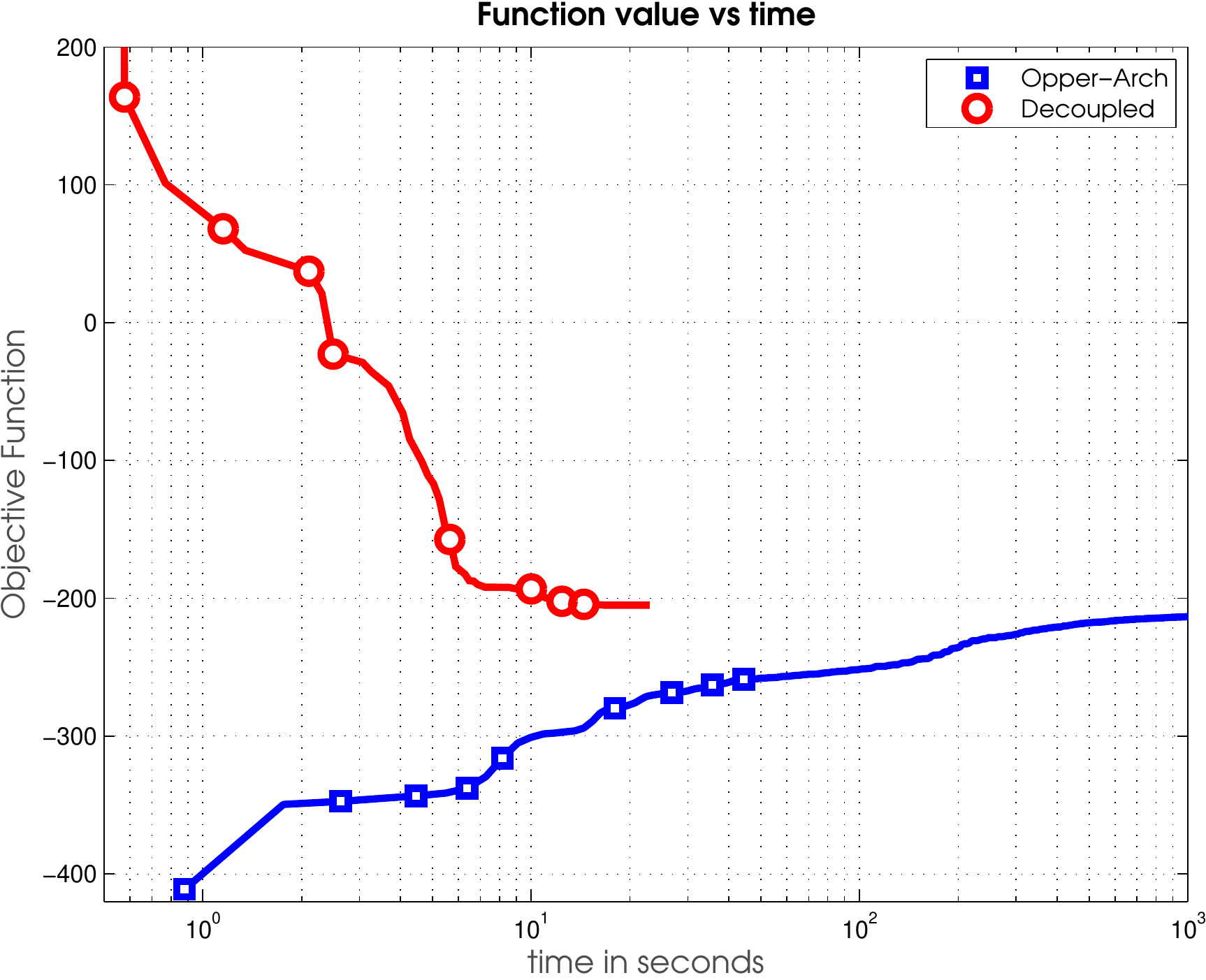}}
\caption{Comparisons for multinomial logit GP classification on the glass dataset. Figure (a) shows the negative log marginal likelihood approximations at the top and prediction errors at the bottom for many values of $\log(s)$ and $\log(\sigma)$. The star shows the minimum of the negative marginal likelihood, which achieves a reasonable prediction error. Figure (b) compares the traces of objective function with respect to time in seconds. We choose the hyperparameter setting which minimizes the negative of train log-likelihood. We see that dual variational inference converges much faster than the existing method.}
\label{fig:multiClassGP}
\end{figure*}

We apply the mGPC model to the forensic glass data set (available from the UCI repository) which has $N=214$ data examples, $K=6$ categories, and features $\vect{x}$ of length 8. 
We use $80\%$ of the dataset for training and the rest for testing.
We set $\vect{\mu}=0$ and use a squared-exponential kernel, for which the $(i,j)$th entry of $\vect{\Sigma}$ is defined as: $\vect{\Sigma}_{ij} = -\sigma^2 \exp[-\half||\vect{x}_i-\vect{x}_j||^2/s]$.
Similar to the setup of \citet{Girolami06}, the prior factorizes across classes and we fix the hyperparameters $\sigma$ and $s$ to be same for all the classes.
We find a good setting of these hyperparameters using the approximate marginal likelihood on training set.
We compute this on a $11\times 11$ grid, giving us total 121 hyperparameter settings.
We also compute the prediction error defined as $-\log_2 \tilde{p}(y_{test}|\vect{\theta},\vect{y}_{train},\vect{x}_{train},\vect{x}_{test})$, where $(\vect{y}_{train},\vect{x}_{train})$ and $(\vect{y}_{test},\vect{x}_{test})$ are training and testing data, respectively. Here, $\tilde{p}(y_{test}|\cdot)$ is the marginal predictive distribution approximated using the Monte Carlo method (see \citet[Chapter 3]{Rasmussen06} for details). 

The results are shown in Fig.~\ref{fig:multiClassGP}(a), where we plot the two quantities.
The star shows the minimum of the negative marginal likelihood.
We see that at this setting the algorithm also achieves a reasonable prediction error.

Fig.~\ref{fig:multiClassGP}(b) shows the traces of the objective function for the two methods.
The Opper-Arch method maximizes the primal objective function while dual variational inference minimizes the dual objective function. 
We show markers for iterations 1, 3, 5, 7, 9, and then at 20, 30, 40, and 50.
We see that the dual inference coverges at least 100 times faster that the existing method (which has not yet converged in the plot).
Each gradient step in Opper-Arch is also more expensive than our method since the number of parameters is $2NK$ (where $K$ is the number of categories) in contrast to our algorithm which require only $NK$ parameters.
In addition, each function evaluation of Opper-Arch is more expensive than ours.
This is due to the additional trace term in the primal problem
\eqp{primal-problem}, which is not present in the dual problem \eqp{decoupledDual}.
Hence, our proposed algorithm has advantage in terms of the rate of convergence, cost of function evaluation, and the number of parameters.

We observed similar trends for other hyperparameter settings.

\begin{figure*}[t]
\centering
\subfigure[]{\includegraphics[height=3in]{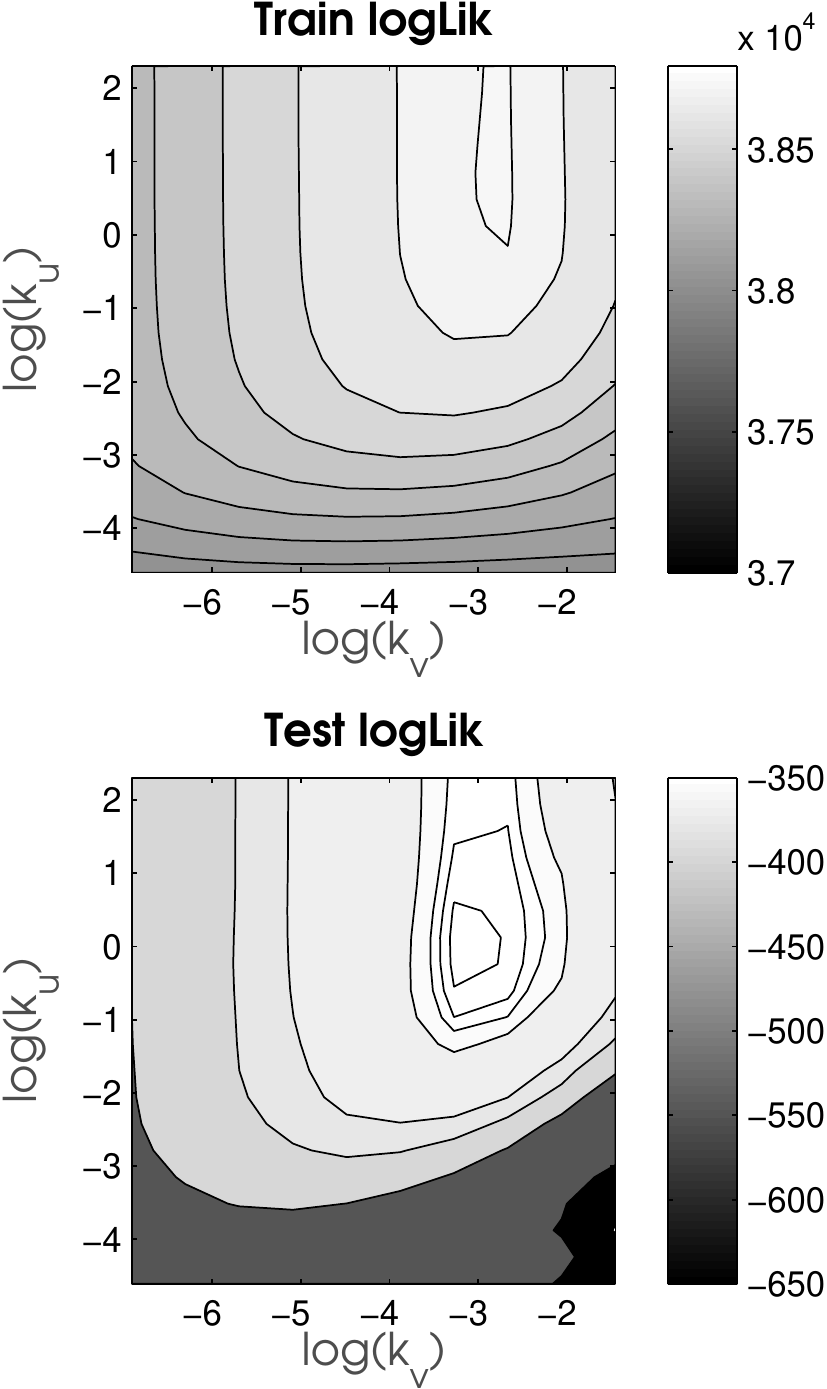}}
\hspace{.2in}
\subfigure[]{\includegraphics[height=3in]{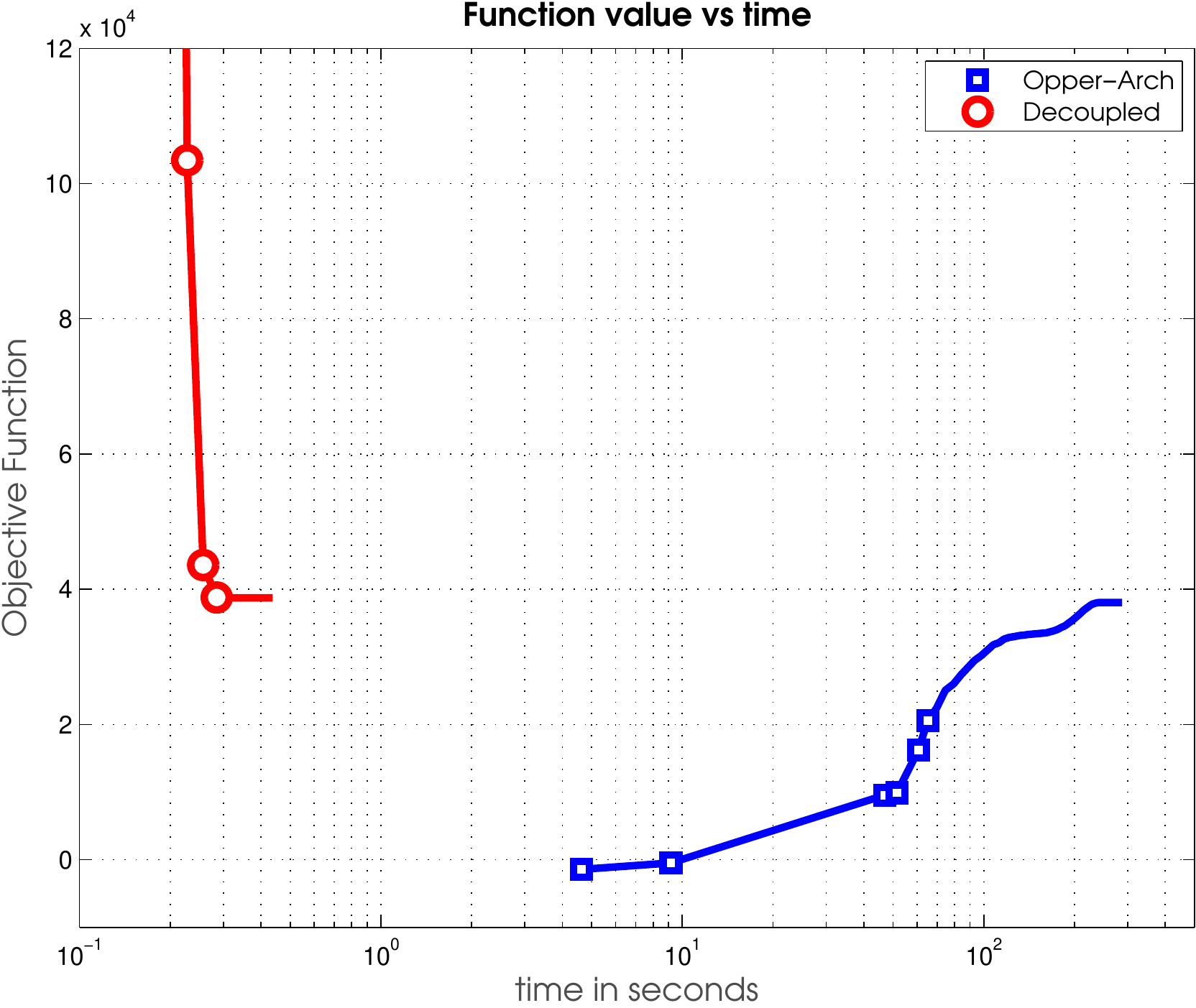}}
\caption{Comparisons for latent GMRF model on the glass dataset. Figure (a) shows the train and test log-likelihood approximations for many values of $k_v$ and $k_u$. Figure (b) compares the traces of objective functions vs time. We choose the hyperparameter setting which maximizes the test log-likelihood.}
\label{fig:spatialAnalysis}
\end{figure*}

\subsection{Latent Gaussian Markov Random Field}
We consider the modeling of the oral cancer mortality rates using a latent GMRF, described in \citet{Rue05}.
The data consists of mortality counts in 544 regions in Germany during 1986-1990.
We model the count $y_i$ in a region $i$ using a Poisson likelihood with the rate $\lambda_i = \exp(\mu + u_i + v_i)$.
Here, $\mu$ is the offset, $v_i$ is an unstructured component, and $u_i$ a spatially structured component.
The prior on the last two terms is shown below in
Eq. \ref{eq:spatialPrior}. We assume an independent Gaussian prior
over $\vect{v}$ with hyperparameter $k_v$, and an intrinsic GMRF of
first-order with hyperparameter $k_u$ (see \citet{Rue05} for details
on GMRFs). Here, $i \sim j$ are all unordered pairs $(i,j)$ such that
regions $i$ and $j$ are neighbors, i.e.,
\begin{align}
p(\vect{u},\vect{v}|k_u,k_v) \propto& \exp\!\Big[{-\half} k_v\sum_i v_i^2 -\half k_u\sum_{i\sim j} (u_i-u_j)^2\Big].
\label{eq:spatialPrior}
\end{align}
The GMRF prior can be easily written in the form of the LGM discussed in Section \ref{sec:model}.

We choose 500 regions at random as training data and keep the rest as testing data. 
For simplicity, we set $\mu$ to 0.
To find a good setting of other hyperparameters, we compute train and test log-likelihoods for several $(k_u,k_v)$.
The results are shown in Fig.~\ref{fig:spatialAnalysis}(a).
We see that the shape of train and test log-likelihoods are similar, justifying the maximization of the train log-likelihood to achieve good test accuracy. 
The maximum occurs at $k_u = 2.637$ and $k_v = 0.088$.

Fig. \ref{fig:spatialAnalysis}(b) shows the traces of optimizers for this setting of $k_u$ and $k_v$. We show markers at iterations 1 to 6.
We see that the proposed algorithm converges just in 6 iterations, and is much faster than the Opper-Arch method. 
Similar to mGPC, our method beats Opper-Arch on the number of iterations to converge, number of parameters, and cost of each function evaluations.


\section{Conclusions} \label{sec:conclus}
We presented a novel dual decomposition viewpoint on the variational Gaussian
inference problem for latent Gaussian models. Our approach applies generally
to any likelihood potential for which convex local variational bounds are
available (e.g., Poisson, Bernoulli-logit, multi-way logistic, super-Gaussian)
and is easy to configure to a new setup. Applying standard optimization
technology to the dual formulation leads to an algorithm which has lower
per iteration cost (time and memory) and can converge in orders of magnitude
less iterations than the previous state of the art.

Dual decomposition has been popular for 
MAP inference in graphical model, for example, see \citet{sontag2011introduction, jojic2010accelerated}.
In this paper, we applied the decomposition to the VG inference problem. 
We would like point that the coordinate-ascent approach of \citet{Khan12nips} also has a dual interpretation.
There, each coordinate update can be interpreted as optimization of an element of the
dual variable (see Appendix of the paper).
Our dual approach improves the approach of \citet{Khan12nips} by allowing parallel updates leading to an efficient 
implementation, while maintaining fast convergence.

A disadvantage of our approach is its restrition to the likelihood potentials
with convex local variational bounds. Extension to the non-convex case remains an open problem
which we would like to research in the future.
We also aim to combine our dual formulation with covariance
decoupling ideas from \citet{Seeger:11}, in order to break the $O(L^3)$
computational barrier and to make variational Gaussian inference applicable
to very large problems.

\subsection*{Appendix}
We describe the maximization with respect to $\vm{}$ and $\mxv{}$ to get \eqp{mstar} and \eqp{Vstar}.
We substitute the definition of $\KL[q(\vz{})\, \|\, p(\vz{})]$ from \eqp{primal-detail} into the Lagrangian \eqp{lagrange}. 
Derivatives of the Lagrangian with respect to $\vm{}$ and $\mxv{}$ are
given by
\begin{align}
&\half(\mxv{}^{-1} - \mxsigma{}^{-1} - \mxw{}^T\diag(\vlam{})\mxw{}) = 0, \\
&-\mxsigma{}^{-1}(\vm{} -\vmu{}) - \mxw{}^T\valpha{} = 0.
\end{align}
Simplifying, we get \eqp{mstar} and \eqp{Vstar}.

\comment{
We give a derivation of \eqp{dual-first}. We ignore terms independent of
$\vm{}$, $\mxv{}$ and denote equality up to an additive constant by
``$\doteq$''. Recall that $q(\vz{})=\gauss(\vm{},\mxv{})$, $p(\vz{}) =
\gauss(\vmu{},\mxsigma{})$, and note that
\[
\begin{split}
  & -\KL[q(\vz{})\, \|\, p(\vz{})] = \Ex_q[\log p(\vz{})] + \Ent[q(\vz{})] \\
  & \doteq \log p(\vm{}) - {\textstyle\frac{1}2}\trace\mxsigma{}^{-1}\mxv{} +
  {\textstyle\frac{1}2}\log|\mxv{}|.
\end{split}
\]
Recall the definition of $\mxa{}$ \eqp{amat}. Adding the Lagrange multiplier
terms, we have
\[
  {\cal L} \doteq \log p(\vm{}) - \valpha{}^T\mxw{}\vm{} -
  {\textstyle\frac{1}2}\trace\mxa{}\mxv{} + {\textstyle\frac{1}2}\log|\mxv{}|.
\]
Here, $\vlam{}^T\diag(\mxw{}\mxv{}\mxw{}^T) =
\trace\mxv{}\mxw{}^T(\diag\vlam{})\mxw{}$. The maximization w.r.t.\ $\mxv{}$
results in
\[
  \mxv{*} = \mxa{}^{-1},\quad {\textstyle\frac{1}2}\trace\mxa{}\mxv{*} +
  {\textstyle\frac{1}2}\log|\mxv{*}| \doteq -{\textstyle\frac{1}2}\log|\mxa{}|.
\]
Moreover, the quadratic function $\log p(\vm{}) -
\valpha{}^T\mxw{}\vm{}$ is maximized by $\vm{*} = \vmu{} -
\mxsigma{}\mxw{}^T\valpha{}$, and
\[
  \log p(\vm{*}) - \valpha{}^T\mxw{}\vm{*} \doteq {\textstyle\frac{1}2}
  \vm{*}^T\mxsigma{}^{-1}\vm{*} \doteq R(\valpha{}).
\]
Altogether, we obtain \eqp{dual-first}.
}

\section*{Acknowledgments} 
MEK and MS were supported by an ERC starting grant (277815-SCALABIM).
MEK would also like to thank Dr. Kevin Murphy for financial support at
the beginning of this project.

\bibliography{icml,papers-short,books}
\bibliographystyle{icml2013}

\end{document}